\documentclass[]{interact}

\usepackage{epstopdf}
\usepackage[caption=false]{subfig}

\usepackage{natbib}

\bibpunct[, ]{(}{)}{;}{a}{}{,}
\renewcommand\bibfont{\fontsize{10}{12}\selectfont}

\theoremstyle{plain}

\theoremstyle{definition}

\theoremstyle{remark}

\usepackage{graphicx}
\usepackage{amsmath,amssymb} 
\usepackage{color}
\usepackage{graphicx}
\usepackage{amsmath, xparse}
\usepackage{booktabs}
\usepackage{subfig}
\usepackage{amsfonts}
\usepackage{xcolor}
\usepackage{amssymb, amsmath}
\usepackage{cite}
\usepackage{float}
\usepackage{svg}

\usepackage{pgfplots}
\pgfplotsset{compat=1.16}

\begin{document}


\title{Regularizing disparity estimation via multi task learning with structured light reconstruction} 

\author{
\name{Alistair Weld \and
Joao Cartucho \and 
Chi Xu \and
Joseph Davids \and
Stamatia Giannarou}
\affil{The Hamlyn Centre, Imperial College London}
}

\maketitle

\begin{abstract}
\textcolor{black}{3D reconstruction is a useful tool for surgical planning and guidance. However, the lack of available medical data stunts research and development in this field, as supervised deep learning methods for accurate disparity estimation rely heavily on large datasets containing ground truth information. Alternative approaches to supervision have been explored, such as self-supervision, which can reduce or remove entirely the need for ground truth. However, no proposed alternatives have demonstrated performance capabilities close to what would be expected from a supervised setup. This work aims to alleviate this issue.}
In this paper, we investigate the learning of structured light projections to enhance the development of direct disparity estimation networks. We show for the first time that it is possible to accurately learn the projection of structured light on a scene, implicitly learning disparity. Secondly, we \textcolor{black}{explore the use of a multi task learning (MTL) framework for the joint training of structured light and disparity. We present results which show that MTL with structured light improves disparity training; without increasing the number of model parameters. Our MTL setup outperformed the single task learning (STL) network in every validation test. Notably, in the medical generalisation test, the STL error was 1.4 times worse than that of the best MTL performance. The benefit of using MTL is emphasised when the training data is limited.} A dataset containing stereoscopic images, disparity maps and structured light projections on medical phantoms and ex vivo tissue was created for evaluation together with virtual scenes. This dataset will be made publicly available in the future.

\end{abstract}


\section{Introduction}
\let\thefootnote\relax\footnote{
Alistair Weld | \email{a.weld20@imperial.ac.uk}\\
This project was supported by UK Research and Innovation (UKRI) Centre for Doctoral Training in AI for Healthcare (EP/S023283/1) and the Royal Society [URF$\setminus$R$\setminus$2 01014] and the NIHR Imperial Biomedical Research Centre}
Recently, it has been shown that when large datasets are available, deep learning approaches define the state-of-the-art in 3D scene reconstruction \citep{Zhao2020MonocularDE, Laga2020ASO}. This is fundamentally due to a neural network's ability to learn more complex representations of image data than can be handcrafted. However, the coupling of data volume and performance is an issue, particularly for domains that have limited data availability such as surgery \citep{Hashimoto2018ArtificialII, Willemink2020PreparingMI}. Capturing large amounts of depth information for surgery, especially Minimally Invasive Surgery (MIS), is laborious, due to issues with hardware constraints and the difficulty of dealing with the tissue; primarily issues with the deformation of the tissue which can disturb information capture. This problems is reflected in the SCARED dataset \citep{Allan2021StereoCA}, which is the largest annotated depth dataset for surgical scenes, but only contains 45 unique and complete depth images. Training on small datasets complicates the development of networks \citep{Qi2020SmallDC, Brigato2021ACL}, due to the risk of overfitting \citep{Ying2019AnOO}. To obviate ground truth, stereo self-supervision \citep{Godard2017UnsupervisedMD} approaches have been developed; learning disparity by training to warp the left image to become the right and vice versa. However, results are still considerably worse than what is achieved by supervised networks \citep{Uhrig2017THREEDV}. \textcolor{black}{Accurate 3D reconstruction can provide surgeons a tool for surgical planning and guidance \citep{Hersh2021AugmentedRI}. Therefore, strategies for overcoming data limitation are desperately needed. Otherwise, the development of accurate neural networks for surgery is unachievable.}

Structured light is currently the most dense and accurate approach for creating ground truth information for depth datasets. Example datasets that were created using structured light include NYU \citep{Silberman2011IndoorSS}, Middlebury Stereo \citep{Scharstein2003HighaccuracySD, Scharstein2014HighResolutionSD} and SCARED. Structured light is the projection of patterns into a scene, which when captured by an imaging camera, allows for depth recovery through analysis of the pattern distortions \citep{Salvi2004PatternCS}. Commonly, these pattern projection images are used exclusively for ground truth depth generation. Once depth has been captured, the pattern projection images are discarded afterwards. However, error will occur in the conversion process due to difficulties relating to the environment, surface properties and the hardware \citep{Rachakonda2019SourcesOE, b824283769d743aeba29c2d3468d1184, 5995321, 1211354}. Primarily, errors will occur at the pixel classification stage which for example can be caused by reflections or hardware malfunction. This means that the information within the structured light images and the generated depth maps are not the same. 

In this paper, we are proposing a unique approach to depth estimation. Firstly, we show that it is possible in itself to teach a neural network to be able to artificially project structured light patterns into a scene. Which is indirectly learning to perform disparity estimation. To the best of our knowledge, this is the first use of a neural network for the learning of the projection of structured light patterns into a scene. Secondly, we show that the dual training of direct disparity regression and structured light projection, via multi task learning, enhances network training and improves disparity learning without increasing the number of parameters nor requiring the collection of additional ground truth data.

\section{Related Work}

\subsection{Active Sensors}

Structured light can be used to produce unique pixel codes for each stereo image, which enables the use of simple matching techniques for the stereo disparity calculation. This approach can provide dense and accurate depth maps for a given stereo image. However, structured light requires a controlled environment, multiple hardware and sometimes temporal variation. Certain structured light patterns require sequential projection of different patterns over time, such as binary/gray code which is what this paper works with. The control of the environment, when performing structured light projection, is important for good pixel classification. Robust classifiers are normally required to prevent incorrect classification \citep{Salvi2004PatternCS}. Theses limitations have large ramification for setups designed for real time performance in dynamic environments. Our proposed method for structured light reconstruction removes the need for the additional hardware and the temporal requirement.

\subsection{Depth estimation}

The state-of-the-art for 3D reconstruction is defined by deep learning approaches that directly regress depth values. The general consensus within the deep learning community is that a neural network will discover a better way to solve the stereo matching problem when given free roam in an end-to-end setup where there is limited human supervision. Examples of these networks could be, RAFTStereo \citep{Lipson2021RAFTStereoMR} or PSMnet \citep{Chang2018PyramidSM} for stereo and DORN \citep{Fu2018DeepOR} for monocular. The prevalent issue with these networks is the requirement of large training datasets. In the papers mentioned, the networks are pre-trained on scene-flow which contains approximately 60k images, then fine tuned on popular sets like KITTI. Although generalisability has been shown to be largely acheivable, the further the data similarity is to what has been trained on, the worse the performance. This poses a large hurdle when the 3D reconstruction task is for specific and niche tasks. In this paper we show that recycling structured light images, which is data commonly acquired to produce depth ground truth, can improve the performance of disparity estimation techniques especially when data is limited.

\subsection{Multi Task Learning}

Multi task learning (MTL) has already been proposed for depth/disparity estimation purposes. In previous works, the training of depth/disparity, has, for example, been combined with the training of semantic segmentation or instance segmentation \citep{sener2018multi, Kendall2018MultitaskLU}. These works have shown that it is possible to achieve an improved depth/disparity performance in comparison to when trained as a single task. However, as has been identified in previous literature \citep{Standley2020WhichTS}, MTL doesn't guarantee improved performance and balancing the combination of tasks and the combination of weights of the loss terms is difficult. In this work we show that it is possible to use structured light reconstruction in a MTL framework to improve disparity learning. Where the ground truth requires no annotation.

\section{Method}

In this section we introduce our stereo disparity estimation techniques. Firstly, we propose the first deep learning model to learn the projection of structured light patterns onto a scene. The goal of this model is to show that it is possible to accurately learn structured light patterns that respect the stereoscopic perspectives and verify that structured light information can be used for 3D reconstruction purposes and should not be discarded after ground truth information has been generated; which is what is most commonly done. Secondly, we propose a novel MTL method for disparity estimation, showing that MTL can enhance disparity learning by dual training on structured light.

\subsection{Dataset generation}

\begin{figure}[t]
\centering
\subfloat[The scene lit with camera light source only.]{\resizebox*{0.35\linewidth}{!}{\includegraphics{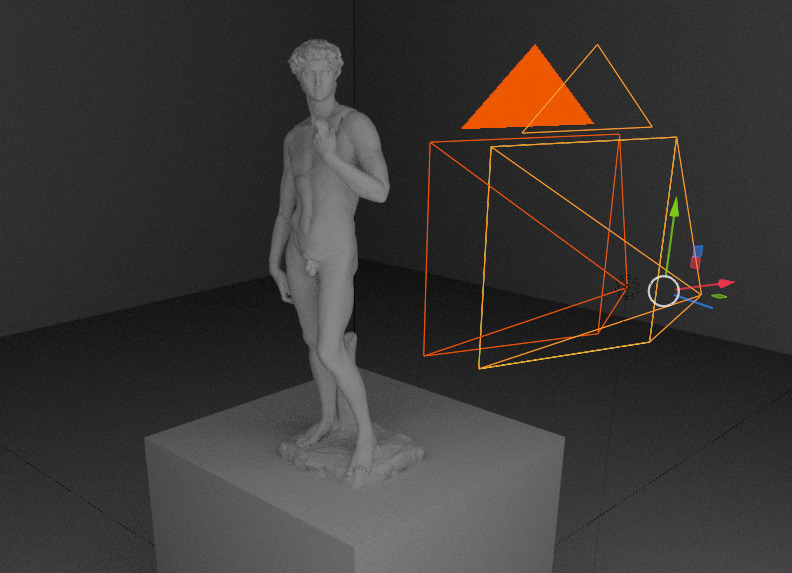}}}\hspace{5pt}
\subfloat[The scene lit with projector only.]{\resizebox*{0.35\linewidth}{!}{\includegraphics{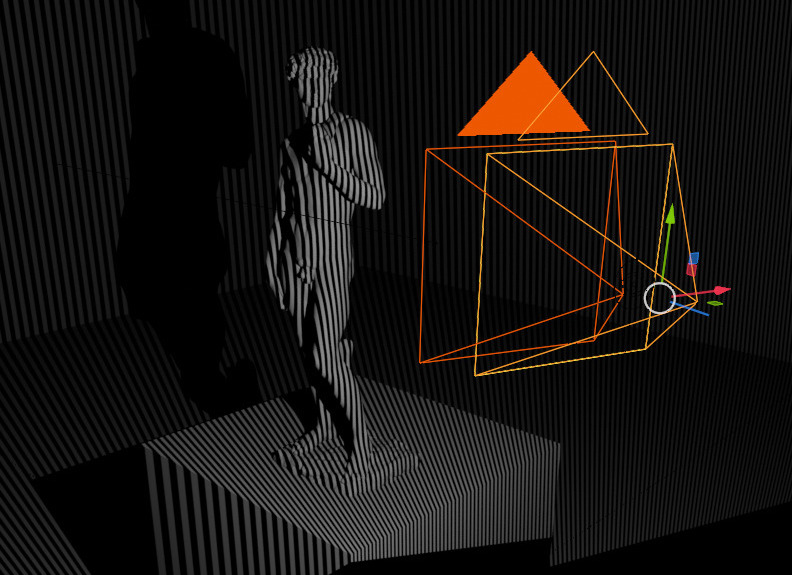}}}\hspace{3pt}
\vspace*{-0.7\baselineskip}
\caption{Demonstration of the virtual environment. The virtual camera is represented in orange and the projector object is represented in yellow.}
\label{im:virtual_scene}
\end{figure}

\noindent\textbf{Virtual dataset:} Due to the lack of any publicly available datasets containing structured light patterns, we had to create our own dataset. A simulated environment was developed using Blender \citep{blender} to enable automatic generation of virtual scenes containing cultural objects, as shown in Fig.\ref{im:virtual_scene}. The VisionBlender add-on \citep{Cartucho2020VisionBlenderAT} was used to generate the ground truth information. 

The virtual environment is simple, consisting of a cubic room of volume 1 $[m^{3}]$ and a central podium of size 10 $[cm^{3}]$, which is used to hold different objects. All the surfaces in the whole virtual scene are grey and there is no texture, including the surfaces of the objects. The projector \citep{Ocupe} was rigidly attached to the camera object with a rotation of $\theta_{y}=1.5^{\circ}$ and a translation of $T=[0.02m, 0.00m, 0.02m]$. 800 cultural artefact objects, were used as the diversity factor in this dataset. The objects contain complex 3D surfaces and a smooth texture, making them challenging to reconstruct using existing 3D reconstruction approaches. All the objects were downloaded from the ``Scan The World'' project which is hosted by MyMiniFactory \citep{scantheworld}. The dimensions were resized using a uniform distribution $U(0.03m,0.1m)$. For each object, 10 stereo image-pairs were captured from different view points. The camera is always looking towards the geometrical centre of the object. The distance between the camera and that geometrical centre is sampled from $U(0.03m,0.1m)$. Respecting these two constraints, the pose of the 10 viewpoints are chosen at random.  

For each viewpoint, the stereo images and the associated ground truth data were generated. The ground truth includes the depth/disparity maps and stereo images without and with 8 projected binary patterns. For each projection, the number of vertical lines equals $2^{n}, \forall n \in [1..8]$, where n is the projection number. The resolution of the captured images are $256\times256$. This dataset contains approximately 8000 images.\\

\begin{figure}[t]
\centering
\subfloat[The capture of a sheep's liver using the dVRK \textcolor{black}{laparoscope}.]{\resizebox*{0.35\linewidth}{!}{\includegraphics[width=0.75\linewidth, height=8cm]{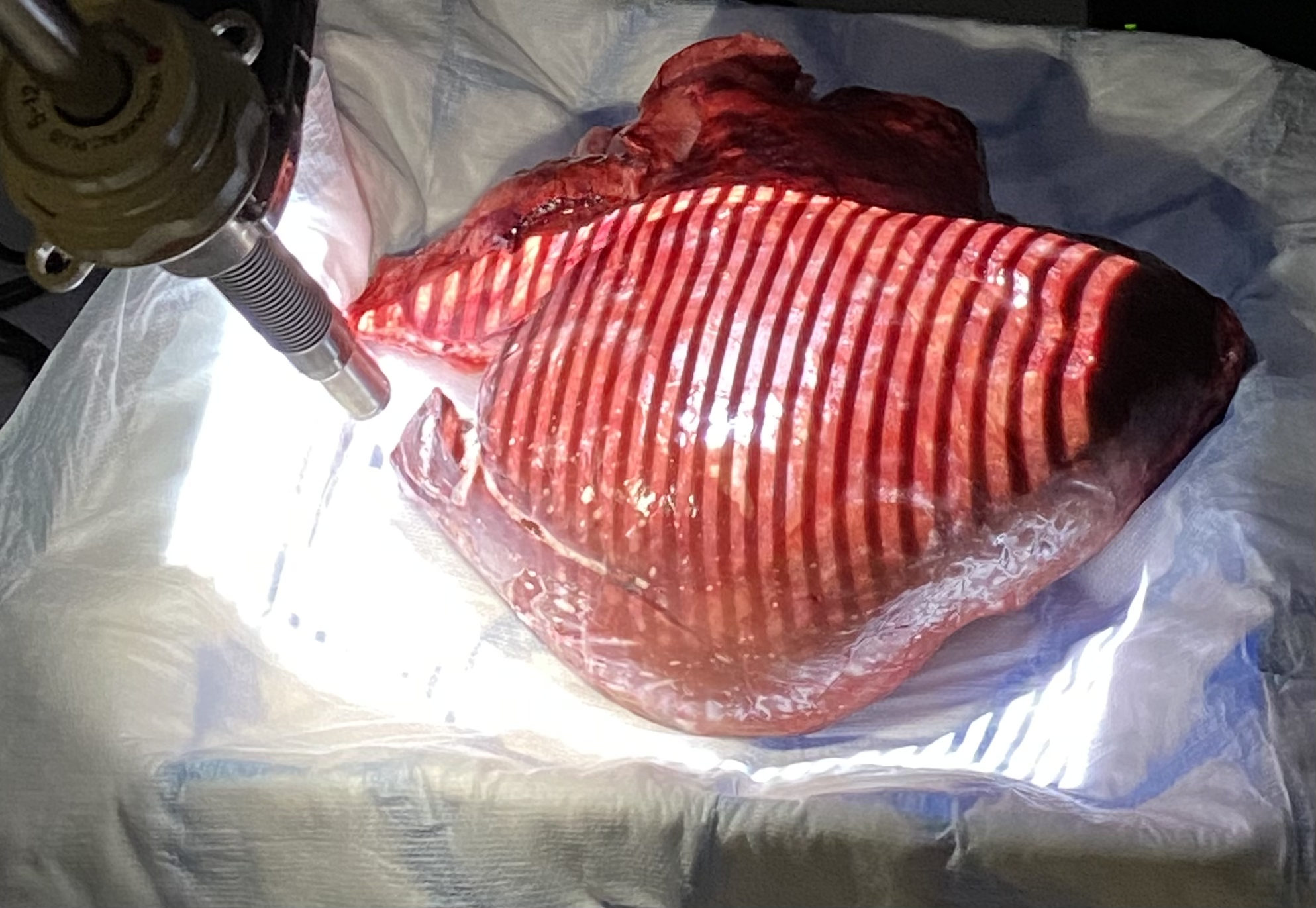}}}\hspace{5pt}
\subfloat[An example image captured which contains multiple ex vivo tissue.]{\resizebox*{0.35\linewidth}{!}{\includegraphics[width=0.75\linewidth, height=8cm]{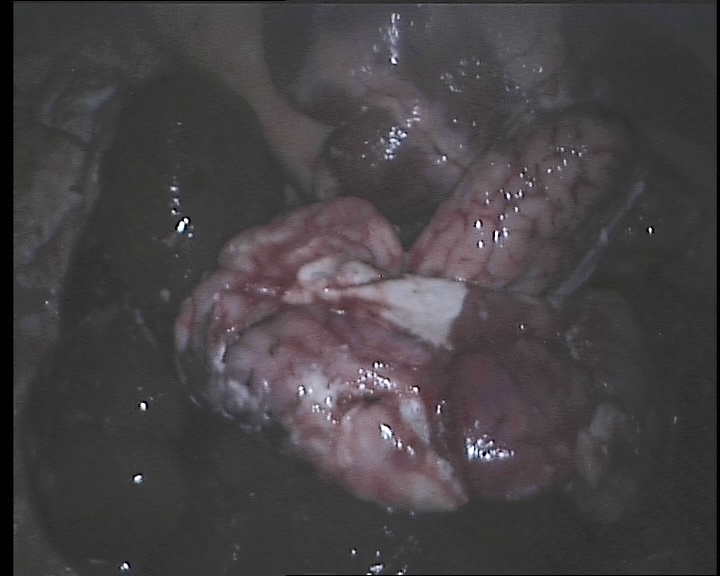}}}\hspace{3pt}
\vspace*{-0.7\baselineskip}
\caption{Shown in (a) and (b) is the setup \textcolor{black}{for the medical dataset collection. Image (a) shows an example of tissue captured using the dVRK laparascope.} (b) does not contain projection patterns and would be fed as an input to the neural network.}

\label{img:paterson}
\end{figure}

 \noindent\textbf{Medical dataset:} A medical dataset was created using the da Vinci Research Kit (dVRK) \citep{kazanzides2014open} to extend the evaluation of the model onto real images. \textcolor{black}{The dVRK was chosen as it is the research model of the surgical system that is used in clinical practice and commonly used for research in this field.} A structured light projector was attached to a da Vinci camera arm and the stereo laparoscope of the machine was used to capture scenes with projected patterns on medical phantoms and ex-vivo organs from sheep, cows and chicken. Gray code was used for this dataset with 9 patterns. The decoding was performed using the three phase algorithm \citep{threephase}. A turning table was used to hold the objects and were rotated for multiple perspectives. Fig. \ref{img:paterson} shows the setup and an example image which would be fed as the input to the neural network (an image without projection patterns).

\subsection{Structured light reconstruction}
\label{Sec:Disparity SLR}

Given a stereo image pair, a neural network has been designed and trained to predict how structured light patterns should project on the scene surfaces of each image, separately. Specifically, the network projects binary and gray code structured light patterns, which consist of vertical bars of white and black colour. To predict these patterns, the UNet \citep{Ronneberger2015UNetCN}  network is trained as a per pixel binary classifier; 1 (white) or 0 (black). These predicted patterns can then be used to generate disparity or depth maps by performing 2D cross correlation over the epipolar lines. This reveals how the network has understood the 3D scene.

The input to the UNet is a pair of rectified stereo images (which do not contain any projection patterns), concatenated along the channel axis, I=[I$_{l}$;I$_{r}$], I$_{l}$,I$_{r}$ $\in$ $\mathbb{R}^{h\times w\times c}$. The network is tasked to predict structured light projections in each input stereo image, where every pixel requires a unique code along a horizontal epipolar line. $P$ denotes the ground truth patterns and $\hat{P}$ the neural network output which is defined as $\hat{P}$ =[$\hat{P}_{l}$;$\hat{P}_{r}$], $\hat{P}_{l}$,$\hat{P}_{r}$ $\in$ $\mathbb{R}^{h\times w\times t} ,0\leq \hat{p} \leq1, p \in P, \hat{p} \in \hat{P}$. The parameter $t$ denotes the number of projected patterns and $t=8$ for the virtual dataset and $t=9$ for the medical dataset. \\

\noindent\textbf{Losses:} Two losses were chosen for the structured light learning. Since the network is trained to produce a binary mask, binary cross entropy (bce) loss defined in Eq.(\ref{eq:sig_cross_entropy}) is used as the primary loss. To emphasise the pattern edges, an L2 loss is used on the horizontal derivatives as in Eq.(\ref{eq:l2}). These derivatives are calculated by convolving over the images using a Prewitt operator mask, generating $D=P*\begin{bmatrix}-1 & 0 & 1\end{bmatrix}$ and $\hat{D}=\hat{P}*\begin{bmatrix}-1 & 0 & 1\end{bmatrix}$ which are the derivatives of the ground truth and predicted output, respectively with elements $ d \in D$; $\hat{d} \in \hat{D}$. These two individual losses are summed together in Eq.(\ref{eq:l3}) with a weighting gain of \textcolor{black}{$\lambda_{1}=1/80$}.

\begin{equation}
\begin{aligned}
\textcolor{black}{\mathcal{L}_{bce} = - \frac{1}{N} \sum_{i=1}^{N} p_{i} \cdot log(\hat{p}) + (1-p_{i}) \cdot log(1-\hat{p}_{i})}
\end{aligned}
\label{eq:sig_cross_entropy}
\end{equation}

\begin{equation}
\begin{aligned}
\textcolor{black}{\mathcal{L}_{L2} = \frac{1}{N} \sum_{i=1}^{N} || d_{i} - \hat{d}_{i} ||} 
\end{aligned}
\label{eq:l2}
\end{equation}

\begin{equation}
\textcolor{black}{L_{sl} = \mathcal{L}_{bce}  + \lambda_{1}\mathcal{L}_{L2}}
\label{eq:l3}
\end{equation}

\noindent\textbf{Extracting disparity from structured light:} To generate the disparity map using the proposed structure light projection network, cross correlation is performed along the epipolar lines of a pair of rectified stereo images, fully connected along the pattern dimension. A patch, represented by $W_{k}$, where k is the reference index, of size $17\times 17\times t$ is taken in $\hat{P}_{l}$, and cross correlation is performed over a range of pixels in $\hat{P}_{r}$ along the epipolar lines. More specifically, for every pixel on the left image, the cross correlation is estimated for pixels on the right image which are along the epipolar line at a distance within the maximum considered disparity for each dataset, individually. In our work, the maximum disparity has been set to $u=25\%$ of the width of the image. The cross correlation on the right image will start from the same position as the examined patch on the left image, striding with a step $s$. The patch size was chosen after tuning. 

\begin{equation}
    \begin{aligned}
        cc(k,s) = \sum_{(x,y) \in W_{k}} \hat{P}_{l}(x,y) \cdot \hat{P}_{r}(x-s,y), \forall s \in [0..u]
    \end{aligned}
\end{equation}

\begin{equation}
disparity = \operatorname*{arg\,max}_{s} cc(k,s)
\end{equation}

\subsection{Direct disparity estimation with multi task leaning}
\label{Sec:Disparity MTL}

For this task, we are exploring the benefits of training disparity and structured light jointly. \textcolor{black}{Specifically, we investigate the relative effect of introducing the MTL framework, whether or not the MTL framework improves the disparity estimation performance produced using STL.} The PSMNet \citep{chang2018pyramid} architecture and training procedure were chosen as it is one of the more recognised architectures for disparity estimation and has recorded competitive performances in benchmark challenges. However, to note, any other architecture could have been used. In this work, the PSMNet architecture was modified in such a way that no added complexity or increased number of parameters is introduced for the disparity estimation path. Compartmentalising the design allows the structured light section to be detached when training is complete. 

\begin{figure}[t]
\centering
\setkeys{Gin}{height=45mm,width=110mm}
{\includegraphics{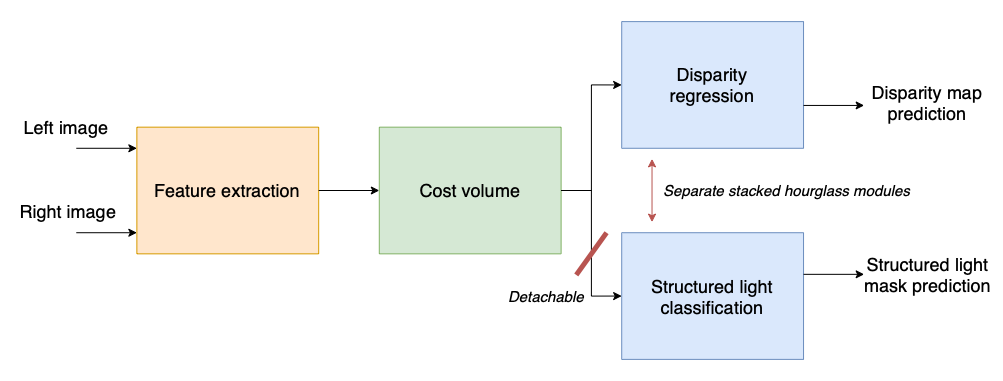}}
\vspace*{-0.7\baselineskip}
\caption{Simple block diagram representation of the multi task learning framework that has been applied on top of the PSMNet architecture.} 
\label{img:arch}
\end{figure}

The PSMNet contains a stacked hourglass module to regress disparity. The modification which we introduced to the architecture to create the multi task framework was done at this point. More specifically, the stacked hourglass module was duplicated; so that there is bifurcation after the cost volume. This results in two parallel paths, one for either task as shown in Fig. \ref{img:arch}. A disparity range of 96 was chosen. The training is performed on RGB images with dimensions $256 \times 256$. In the evaluation stage, the outputs are resized using bilinear interpolation to match the original dimensions which are $256\times256$ for the virtual images, $720\times576$ for the medical images  and $1280\times1024$ for the SCARED images.

The loss function is composed of the structured light cross entropy loss Eq.(\ref{eq:sig_cross_entropy}) and an L2 disparity loss. In our work, three different weighting strategies have been used to combine these loss terms. This is because multi task learning is complex and different strategies benefit different tasks in a way which is generally unknown before experimentation is performed. Firstly, a simple constant scaling of the structured light loss was done with Eq.(\ref{eq:loss1}). Secondly, the training strategy from \citep{liu2019end} was implemented Eq.(\ref{eq:loss2})-(\ref{eq:loss3}), which modifies the weights every epoch. \textcolor{black}{Parameters $\lambda_{3}$ and $\lambda_{4}$ are the task weightings; a product of the ratio of the previous epoch's total losses.} Here, $\zeta_{1}=2$ and $\zeta_{2}=2$. \textcolor{black}{The parameter $\zeta_{2}$ is used to balance the task weighting distribution, and $\zeta_{1}$ is used to gain the softmax. The values were chosen after extensive experimentation.} Thirdly, the uncertainty weighting strategy from \citep{Kendall2018MultitaskLU} was implemented as in Eq.(\ref{eq:loss4}). The weight \textcolor{black}{$\lambda_{5}=0.5$} has been introduced to prioritise the disparity learning. The $\sigma$ parameters denote the observed noise\textcolor{black}{, in practice, these are learnable parameters used to weight each loss.}

\begin{equation}
\textcolor{black}{L_{const} = \lambda_{2} \mathcal{L}_{sl} + \mathcal{L}_{disp}}
\label{eq:loss1}
\end{equation}

\begin{equation}
\textcolor{black}{\lambda_{k} = \zeta_{1}\frac{exp(\omega_k \zeta_{2})}{\sum_{i \in \{3,4\}}\omega_i \zeta_2}, \omega_{k}=\frac{\mathcal{L}_{task}(t-1)}{\mathcal{L}_{task}(t-2)}, k \in [3,4], task =  \begin{cases}
      sl, & \text{if}\ k=3 \\
      disp, & \text{if}\ k=4
    \end{cases}}
\label{eq:loss2}
\end{equation}

\begin{equation}
\textcolor{black}{L_{epr} = \lambda_{3} \mathcal{L}_{sl} + \lambda_{4} \mathcal{L}_{disp}}
\label{eq:loss3}
\end{equation}

\begin{equation}
\textcolor{black}{L_{unc} = \frac{\lambda_{5}}{2 \sigma_{sl}^2} \mathcal{L}_{sl} + \frac{1}{2 \sigma_{disp}^2} \mathcal{L}_{disp} + log(\sigma_{sl}) + log(\sigma_{disp})}
\label{eq:loss4}
\end{equation}

\section{Results}

\newlength\colomnonelength
\setlength\colomnonelength{\dimexpr.20\columnwidth-3\tabcolsep-0.25\arrayrulewidth\relax}
\newlength\colomnonelengthtwo
\setlength\colomnonelengthtwo{\dimexpr.2\columnwidth-3\tabcolsep-0.25\arrayrulewidth\relax}
\newlength\mylength
\setlength\mylength{\dimexpr.15\columnwidth-3\tabcolsep-0.25\arrayrulewidth\relax}

\subsection{Evaluation of structured light reconstruction}

In this section, the reconstruction capability of the proposed artificially projected structured light model is assessed, using the disparity maps generated with cross correlation, as described in Sec. \ref{Sec:Disparity SLR}. The aim of this validation is not to compare the performance of the proposed structured light reconstruction model to state-of-the-art disparity estimation models. Rather, we want to explore the benefits of using this alternative approach for disparity estimation and evaluate whether or not the results from the proposed method meet the expected levels of accuracy produced by the conventional approach of directly regressing disparity. \textcolor{black}{More specifically, the hypothesis that this validation aims to verify is that using the structured light in the training process, through Multi-Task Learning (MTL), offers improved performance compared to Single-Task Learning (STL). Without requiring extra parameters for the disparity estimation path and not requiring extra data collection; alleviating issues of data limitation.}

A comparison network was trained to directly estimate disparity
; using the same UNet architecture. The network was trained to predict disparity using a scaled Sigmoid. The output is mapped to the width of the image. An L2 on the disparities was used as the loss function. This network was used to control the experiments and remove the influence of the network architecture on the comparison of the results. There are no major architectural differences between the network proposed for structured light reconstruction and this comparison network. With the exception of the final layer which has been modified to accommodate the output dimension requirements. Comparison to state-of-the-art disparity estimation models is out of the scope of this validation as our focus is to prove the viability of reconstructing structure light.

The 3D reconstruction results are presented in Table \ref{tab:mae_main}. The first and second column contain the mean absolute error (MAE) for each model. Both training and evaluation is performed on grayscale images of dimension $256 \times 256$. The network was trained on a single NVIDIA GeForce RTX 3080 10GB graphics card. Adam \citep{Kingma2015AdamAM} was chosen as the optimizer, with a fixed learning rate of 0.0001. The deep learning model was coded using the PyTorch framework \citep{NEURIPS2019_9015}  \\

\begin{figure}[t]
\centering
\setkeys{Gin}{height=27mm,width=30mm}
    \subfloat[True left.]{\includegraphics{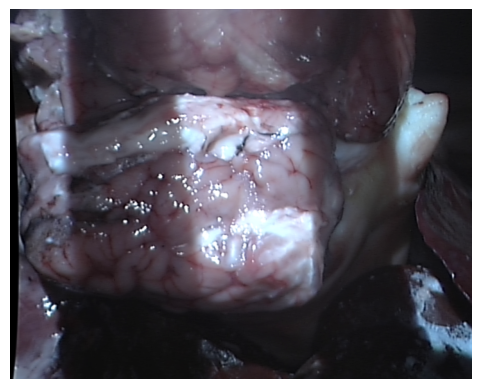}}
    \subfloat[True Right.]{%
        \includegraphics{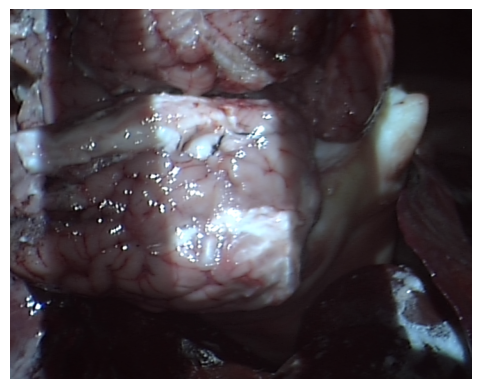}}
    \subfloat[True disparity.]{\includegraphics{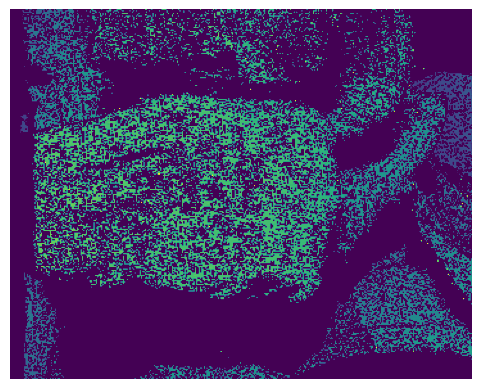}} \\
    \subfloat[Predicted left.]{%
        \includegraphics{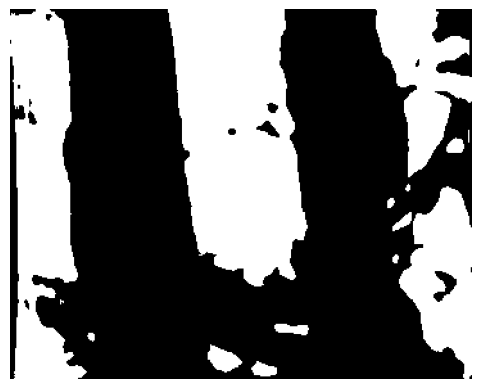}}
    \subfloat[Predicted right.]{\includegraphics{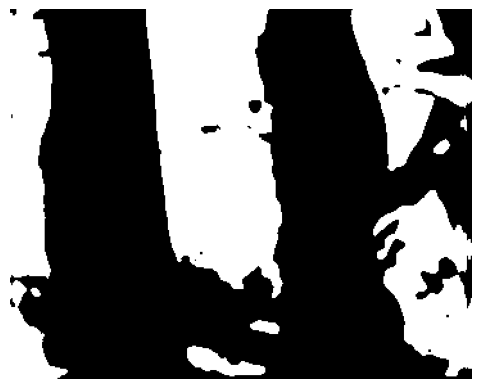}}
    \subfloat[Predicted disparity.]{%
        \includegraphics{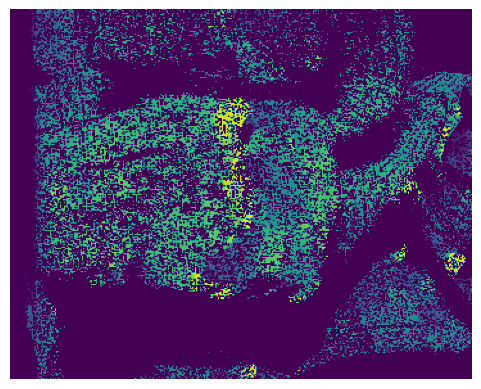}}
    \vspace*{-0.7\baselineskip}
    \caption{This figure contains an example output from the structured light network applied on the medical dataset. The top row contains the ground truth images and the bottom row shows the predictions.}
\label{img:medical}
\end{figure}

\noindent\textbf{Virtual dataset validation:} The results on the virtual dataset, at the first two rows of Table \ref{tab:mae_main}, show that it is possible to achieve good reconstruction using the proposed approach. Virtual Seg denotes the metrics for the virtual object and the podium, segmented using masks provided by VisionBlender. Comparison between both approaches shows that comparable performance can be expected. For the virtual data, the training provided to the disparity network is higher quality than the structured light, as the disparity is extracted directly from the simulation environment. We hypothesise that this is the primary cause of the performance difference between the two implementations. \textcolor{black}{The training/test split was 80/20, at the object level.} \\

\noindent\textbf{Medical dataset validation:} Taking the networks trained on the virtual scenes, fine tuning on the medical dataset, shown in Fig. \ref{img:medical}, is performed to reveal the benefit of using structured light when the size of the datasets are limited. The training data for the structured light was 9 times larger than for the direct disparity network because 9 patterns were collected for every depth map. The results are shown in the bottom row of Table \ref{tab:mae_main}. The accuracy recorded from the proposed method was twice as high as the direct regression approach. This verifies that having the larger volume of data for structured light training, when the datasets are small, improves the accuracy of the estimated disparity because of the increased complexity of the task and the increased number of training samples. \textcolor{black}{The training/test split was 80/20, at the keyframe level.}\\

\newlength\mylengthbbb
\setlength\mylengthbbb{\dimexpr.28\columnwidth-3\tabcolsep-0.25\arrayrulewidth\relax}
\begin{table}[t]
  \footnotesize
  \begin{center}
    \caption{Mean absolute error (MAE) produced by the proposed structured light network (with the disparity produced using the cross correlation) and the direct disparity network. }
    \label{tab:mae_main}
    \newcommand{\fl}[1]{\multicolumn{1}{C}{#1}}
    \begin{tabular}{p{\colomnonelength}|p{\mylengthbbb}|p{\mylengthbbb}}
      \toprule 
      \textbf{Dataset}
      & \multicolumn{2}{c}{\textbf{Models \tiny{MAE ($pixels$)}}} \\
      \cmidrule(l){2-3}
      & \textbf{Structured light} 
      & \textbf{Direct disparity} \\
      \midrule 
      Virtual  & 1.72 & \textbf{0.52}\\
      Virtual Seg  & 1.45 & \textbf{0.75} \\
      Medical  & \textbf{2.36} & 5.89 \\
      \bottomrule 
    \end{tabular}
  \end{center}
\end{table}

\begin{figure}[t]
\centering
\setkeys{Gin}{height=35mm,width=35mm}
    \subfloat[Disparity error.]{\includegraphics{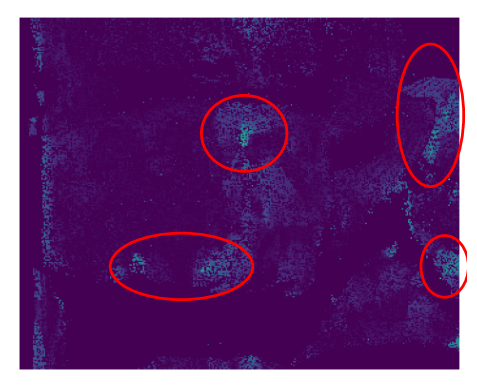}}
    \subfloat[Left uncertainty.]{%
        \includegraphics{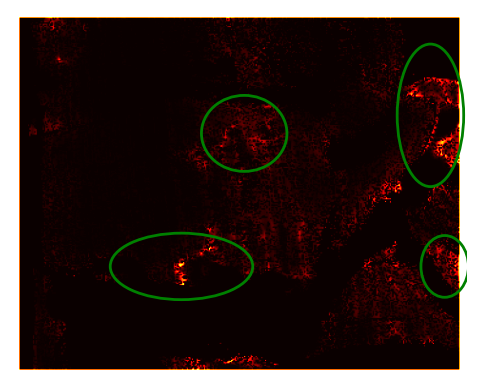}}
    \subfloat[Right uncertainty.]{\includegraphics{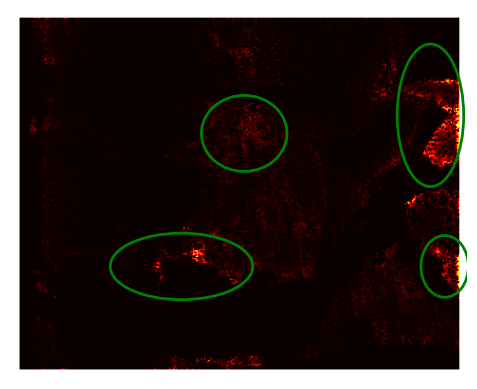}}
    \vspace*{-0.7\baselineskip}
    \caption{Disparity accuracy and uncertainty correlation. The brighter the pixel, the greater the uncertainty and the less accurate. The ellipses are drawn over the same pixels in all three images. Correlating high energy is observed in the areas within the ellipses.} 
\label{img:confidence}
\end{figure}

\begin{figure}[t]
\centering
\setkeys{Gin}{height=45mm,width=55mm}
    {\includegraphics{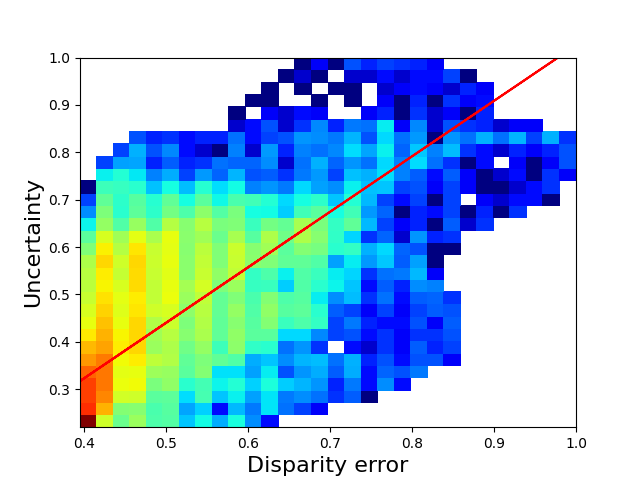}}
    \vspace*{-0.7\baselineskip}
    \caption{Log scaled correlation between uncertainty and prediction error. With a line of best fit drawn, in red, highlighting the positive correlation. The colourmap is set to 'jet'. The x and y axis are both normalised.} 
\label{img:confidence_hist}
\end{figure}

\noindent\textbf{Uncertainty:} 
As the network performs classification, it is easy to acquire the confidence in the predictions \citep{Cao2018EstimatingDF}; the closer the Sigmoid output is to 0.5 the less confident the network is. An empirical assessment of the correlation between the disparity error and network confidence was made. It was concluded that there was a correlation between confidence and accuracy. An example of this is shown in Fig. \ref{img:confidence}. The ellipses highlight the same areas in each image. These results show a correlation between areas of high uncertainty and areas of low accuracy. Fig. \ref{img:confidence_hist} displays a histogram plot of the uncertainty and error in the above example. Each histogram bin represents the number of occurrences for each combination of uncertainty and disparity error. A positive correlation is seen between the disparity error and the uncertainty. This is numerical evidence of the correlation observed in Fig. \ref{img:confidence}. Analysing confidence is critical for healthcare application and the adoption of deep learning methods in clinical practice. This is an advantage for using structured light reconstruction, over disparity estimation, as this is naturally a classification task (where the classification is the final output).

\subsection{Evaluation of the benefit of using multi task learning for the enhancement of disparity learning}

In this section, we explore the benefit of using joint training for the purpose of improving the performance of the disparity estimation compared to when performing single task learning. 5 algorithms are explored here. The standard PSMNet for single task learning (disparity) is used as the benchmark; denoted by $stl$. Then the modified, multi task learning PSMNet is trained 4 times using 4 different strategies. Firstly, $L_{const}$ in Eq.(\ref{eq:loss1}) is implemented twice, where $\lambda_{1} = 0.5, 10$\textcolor{black}{; empirically chosen.} For simplicity we denote the constant gain algorithms using: $cg_{a}:=(\lambda_{1}=10)$ and $cg_{b}:=(\lambda_{1}=0.5)$. Secondly, $L_{epr}$ in Eq.(\ref{eq:loss3}) is implemented; denoted by $epr$. Finally, $L_{unc}$ in Eq.(\ref{eq:loss4}) is implemented; denoted by $unc$. \\

\noindent\textbf{Virtual dataset validation:} 
Firstly, all algorithms are assessed using the virtual dataset. The results are shown in Table \ref{tab:mae_mtl_vd}. Again, Full is the MAE for the entire scene and Seg is the MAE for the podium and object only. This experiment is performed twice, firstly on the entire training set and secondly, on 1/16th of the training set; to see the impact of learning on a smaller dataset. 120 epochs were used for training; we determined this number after multiple experiments, balancing computation time and convergence. For the entire dataset experiment, all networks perform quite similarly, but with two of the multi task learning algorithms achieving slightly better results. However, on the 1/16th experiment the relative performance of $stl$ drops, which highlights the regularizing benefits of using the proposed MTL framework. The extra complexity and the extra data has prevented overfitting to the fewer training samples. \textcolor{black}{We observe that $cg_{b}$ also drops in ranking; highlighting the difficulty of balancing the weighting distribution.} \\

\newlength\colomnonelengtha
\setlength\colomnonelengtha{\dimexpr.125\columnwidth-3\tabcolsep-0.25\arrayrulewidth\relax}
\newlength\mylengtha
\setlength\mylengtha{\dimexpr.23\columnwidth-3\tabcolsep-0.25\arrayrulewidth\relax}
\begin{table}[t]
  \footnotesize
  \begin{center}
    \caption{Comparison of multi task learning vs single task learning on the virtual dataset. Two types of experiments were run, the first was using the full training set, the second was using 1/16th of the training set. Assessing the general performance of each algorithm when training on a large and a small dataset.}
    \label{tab:mae_mtl_vd}
    \newcommand{\fl}[1]{\multicolumn{1}{C}{#1}}
    \begin{tabular}{p{\colomnonelengtha}|p{\mylengtha}|p{\mylengtha}|p{\mylengtha}|p{\mylengtha}}
      \toprule 
      \textbf{Rank} 
      & \multicolumn{2}{c}{\textbf{Full training dataset}} 
      & \multicolumn{2}{c}{\textbf{1/16th training dataset}} \\ 
      \cmidrule(l){2-3} \cmidrule(l){4-5} 
      & \textbf{Full \tiny MAE ($pixels$)} & \textbf{Seg \tiny MAE ($pixels$)} & \textbf{Full \tiny MAE ($pixels$)} & \textbf{Seg \tiny MAE ($pixels$)}  \\

      \midrule 
      1st  & $cg_{b}$ : 0.203     & $cg_{b}$ : 0.380  & $cg_{a}$ : 0.496     & $cg_{a}$ : 0.745   \\[1ex]
      2nd  & unc : 0.207    & unc : 0.381      & $cg_{b}$ : 0.504    & unc : 0.778      \\[1ex]
      3rd  & \textbf{stl : 0.208}            & \textbf{stl : 0.382}      & unc : 0.516            & epr : 0.781       \\[1ex]
      4th  & epr : 0.217            & epr : 0.384     & epr : 0.522            & \textbf{stl : 0.790}         \\[1ex]
      5th  & $cg_{a}$ : 0.217            & $cg_{a}$ : 0.389  & \textbf{stl : 0.549}            & $cg_{b}$ : 0.808   \\[1ex]
      \bottomrule 
    \end{tabular}
  \end{center}
\end{table}

\noindent\textbf{Generalisability evaluation:} 
We explore the generalisability properties of the networks trained on the entire virtual dataset by evaluating them on the SCARED dataset \textcolor{black}{and the dense Middlebury 2014 Stereo training dataset. The virtual dataset that was used for training is limited, and therefore, benchmark results were not expected to be achieved. This test was constructed only to compare the performance of using STL and MTL; investigating the impact of the MTL framework with structured light learning.} The SCARED dataset contained two test sets \textcolor{black}{(TS8 and TS9)}, which collectively contain 45 unique images, which were warped to create a much larger volume of test data. So, to avoid the influence of error created during the artificial warping, only the 45 unmodified and complete images are used for evaluation. Table \ref{tab:generalise} shows the results of this experiment. The main point of interest here is the position of the $stl$ algorithm. In \textcolor{black}{all datasets}, the $stl$ algorithm performs worst. What can be understood from these results is that even though the performance of $stl$ was comparable to the multi task learning performances on the virtual dataset, it has overfitted \textcolor{black}{to the virtual dataset distribution}. Whereas, the MTL framework has provided a regularizing effect during training, which has resulted in greater generalisability.  \\

\newlength\colomnonelengthaa
\setlength\colomnonelengthaa{\dimexpr.097\columnwidth-3\tabcolsep-0.25\arrayrulewidth\relax}
\newlength\mylengthaa
\setlength\mylengthaa{\dimexpr.166\columnwidth-3\tabcolsep-0.25\arrayrulewidth\relax}
\begin{table}[t]
\textcolor{black}{
  \footnotesize
  \begin{center}
    \caption{Generalisability evaluation using SCARED \citep{Allan2021StereoCA} and the Middlebury 2014 \citep{Scharstein2014HighResolutionSD}. The choice of displaying the depth error for SCARED and disparity error for Middlebury reflects what is commonly done. The results show that the $stl$ model has achieved poorer generalisability than the MTL approaches. TS denotes the test sets in SCARED. IQR denotes the interquartile range}
    \label{tab:generalise}
    \newcommand{\fl}[1]{\multicolumn{1}{C}{#1}}
    \begin{tabular}{p{\colomnonelengthaa}|p{\mylengthaa}|p{\mylengthaa}|p{\mylengthaa}|p{\mylengthaa}|p{\mylengthaa}|p{\mylengthaa}}
      \toprule 
      \textbf{Rank} 
      & \multicolumn{2}{c}{\textbf{SCARED - TS 8}} 
      & \multicolumn{2}{c}{\textbf{SCARED - TS 9}} 
      & \multicolumn{2}{c}{\textbf{Middlebury}} \\ 
      \cmidrule(l){2-3} \cmidrule(l){4-5} \cmidrule(l){6-7} 
      & \textbf{Depth \newline \tiny MAE ($mm$)} & \textbf{Depth \newline \tiny IQR}  & \textbf{Depth \newline \tiny MAE ($mm$)} & \textbf{Depth \newline \tiny IQR} & \textbf{Disparity \newline \tiny MAE ($pixels$)} & \textbf{Disparity \newline \tiny IQR} \\
      \midrule 
      1st & unc : 25.15 & unc : 21.71 & unc : 27.86 & unc : 24.63     & unc : 58.41 & unc : 76.17\\[1ex]
      2nd  & $cg_{a}$ : 29.01 & $cg_{a}$ : 27.94 & $cg_{a}$ : 32.41 & $cg_{a}$ : 28.10    & epr : 61.43 & epr : 76.20 \\[1ex]
      3rd & $cg_{b}$ : 29.96 & $cg_{b}$ : 31.36 & epr : 31.92 & $cg_{b}$ : 31.07  & $cg_{b}$ : 63.27 & $cg_{b}$ : 81.48\\[1ex]
      4th & epr : 31.38 & epr : 35.43 & $cg_{b}$ : 34.62 & epr : 31.70 & $cg_{a}$ : 64.89 & $cg_{a}$ : 83.30\\[1ex]
      5th & \textbf{stl : 42.10} & \textbf{stl : 73.78} & \textbf{stl : 51.07} & \textbf{stl : 78.83} & \textbf{stl : 75.06} & \textbf{stl : 94.76} \\[1ex]
      \bottomrule 
    \end{tabular}
  \end{center}
}
\end{table}

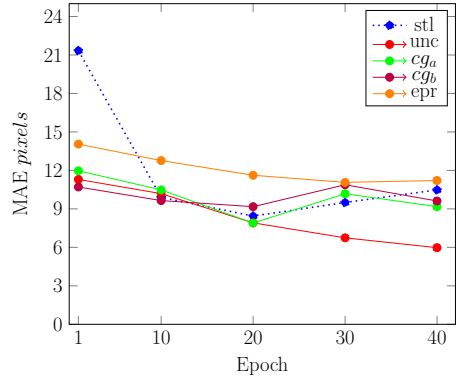
\begin{figure}[!htbp]
\centering
\resizebox{0.43\textwidth}{!}{
\begin{tikzpicture}[thick,scale=1., every node/.style={scale=0.6}]
\LARGE

\begin{axis}[
    xlabel=Epoch,
    ylabel= MAE $pixels$,
    xmin=0, xmax=42,
    ymin=0, ymax=25,
    xtick={1, 10, 20, 30, 40},
    ytick={0, 3,...,25}
            ]
\addplot[dotted, thick,mark=*,blue] plot coordinates {
    (1,21.34)
    (10,9.902)
    (20,8.441)
    (30,9.497)
    (40,10.48)
};
\addlegendentry{stl}

\addplot[->,mark=*,red] plot coordinates {
    (1,11.296)
    (10,10.189)
    (20,7.912)
    (30,6.742)
    (40,5.975)
};
\addlegendentry{unc}

\addplot[->,mark=*,green] plot coordinates {
    (1,11.971)
    (10,10.48)
    (20,7.897)
    (30,10.181)
    (40,9.175)
};
\addlegendentry{$cg_{a}$}

\addplot[->,mark=*,purple] plot coordinates {
    (1,10.708)
    (10,9.648)
    (20,9.172)
    (30,10.881)
    (40,9.616)
};
\addlegendentry{$cg_{b}$}
\addplot[->,mark=*,orange] plot coordinates {
    (1,14.05)
    (10,12.765)
    (20,11.622)
    (30,11.066)
    (40,11.21)
};
\addlegendentry{epr}
\end{axis}
\end{tikzpicture}
}
\vspace*{-0.7\baselineskip}
\caption{The blue dotted line is the performance of the $stl$ model, which is used as a benchmark. What can be seen is that the only stably training algorithm is $unc$, in red, which also produces the best accuracy. All other models begin to overfit after 20 epochs.}
\label{fig:graph}
\end{figure}

\noindent\textbf{Medical dataset validation:} 
Commonly in medical imaging, training data with ground truth is limited. To tackle this limitation, the standard strategy is to pre train a neural network on general and large datasets and then fine tune on data specific to the task. To replicate this scenario, we use the models developed on the entire virtual dataset and fine tune on the medical dataset that we created. Each network was fine tuned for 40 epochs. The results shown in Fig. \ref{fig:graph} highlight the training stability of the compared models. The blue dotted line represents the $stl$ performance. Overfitting can be inferred from the descent behaviour \textcolor{black}{across the 40 epochs,} due to the divergence of the validation accuracy. This is also reflected in the performance for the MTL models: $epr$, $cg_{a}$, $cg_{b}$. However, for $unc$ the training is stable and it also achieves the best \textcolor{black}{disparity} MAE performance. Which again demonstrates the benefit of the multi task learning framework, when the correct training strategy is implemented. 

\section{Conclusion}

In this paper, a novel approach to solving disparity estimation has been proposed that uniquely uses structured light information. We have proposed the first neural network to artificially project structured light patterns onto stereo images. This has allowed 3D reconstruction to be achieved using simple post processing similarity metrics. The proposed 3D reconstruction approach requires no explicit depth information during training.
The performance evaluation results show that the proposed model accurately respects the surface geometry and achieves similar performance when compared to a direct regression network. As this proposed approach uses classification, it is also possible to estimate confidence in the disparity predictions, which is critical for tasks with high risk. This research was then extended, by designing a novel MTL framework to jointly predict structured light and disparity. Our validation verifies that introducing this MTL framework improves the generalisability and capability of learning from small datasets, for disparity estimation. All without increasing the number of parameters for the disparity estimation and using data which is already available and does not require extra annotation. Specifically, the MTL model $unc$ produced results that were consistently better than $stl$, which demonstrates that when the correct multi task learning strategy is implemented, this is a better approach for developing a direct disparity estimation network. Our future work will focus on expanding our database to allow further validation of our work.

\bibliographystyle{tfcse}
\bibliography{interactcsesample}

\begin{thebibliography}{37}
\providecommand{\natexlab}[1]{#1}
\providecommand{\url}[1]{\normalfont{#1}}
\providecommand{\urlprefix}{Available from: }

\bibitem[Allan et~al.(2021)]{Allan2021StereoCA}
Allan~M, McLeod~AJ, Wang~CC, Rosenthal~JC, Hu~Z, Gard~N, Eisert~P, Fu~K,
  Zeffiro~T, Xia~W, et~al. 2021. Stereo correspondence and reconstruction of
  endoscopic data challenge. ArXiv. abs/2101.01133.

\bibitem[Beck(2019)]{scantheworld}
Beck~J. 2019. Scan the world;
  [\url{https://www.myminifactory.com/scantheworld}].

\bibitem[Brigato and Iocchi(2021)]{Brigato2021ACL}
Brigato~L, Iocchi~L. 2021. A close look at deep learning with small data. 2020
  25th International Conference on Pattern Recognition (ICPR):2490--2497.

\bibitem[Cao et~al.(2018)]{Cao2018EstimatingDF}
Cao~Y, Wu~Z, Shen~C. 2018. Estimating depth from monocular images as
  classification using deep fully convolutional residual networks. IEEE
  Transactions on Circuits and Systems for Video Technology. 28:3174--3182.

\bibitem[Cartucho et~al.(2020)]{Cartucho2020VisionBlenderAT}
Cartucho~J, Tukra~S, Li~Y, Elson~DS, Giannarou~S. 2020. Visionblender: a tool
  to efficiently generate computer vision datasets for robotic surgery.
  Computer Methods in Biomechanics and Biomedical Engineering: Imaging \&
  Visualization. 9:331 -- 338.

\bibitem[Chang and Chen(2018{\natexlab{a}})]{Chang2018PyramidSM}
Chang~JR, Chen~YS. 2018{\natexlab{a}}. Pyramid stereo matching network. 2018
  IEEE/CVF Conference on Computer Vision and Pattern Recognition:5410--5418.

\bibitem[Chang and Chen(2018{\natexlab{b}})]{chang2018pyramid}
Chang~JR, Chen~YS. 2018{\natexlab{b}}. Pyramid stereo matching network. In:
  Proceedings of the IEEE conference on computer vision and pattern
  recognition. p. 5410--5418.

\bibitem[Community(2018)]{blender}
Community~BO. 2018. Blender - a 3d modelling and rendering package. Stichting
  Blender Foundation, Amsterdam: Blender Foundation.
  \urlprefix\url{http://www.blender.org}.

\bibitem[Fu et~al.(2018)]{Fu2018DeepOR}
Fu~H, Gong~M, Wang~C, Batmanghelich~K, Tao~D. 2018. Deep ordinal regression
  network for monocular depth estimation. 2018 IEEE/CVF Conference on Computer
  Vision and Pattern Recognition:2002--2011.

\bibitem[Godard et~al.(2017)]{Godard2017UnsupervisedMD}
Godard~C, Aodha~OM, Brostow~GJ. 2017. Unsupervised monocular depth estimation
  with left-right consistency. 2017 IEEE Conference on Computer Vision and
  Pattern Recognition (CVPR):6602--6611.

\bibitem[Gupta et~al.(2011)]{5995321}
Gupta~M, Agrawal~A, Veeraraghavan~A, Narasimhan~SG. 2011. Structured light 3d
  scanning in the presence of global illumination. In: CVPR 2011. p. 713--720.

\bibitem[Hashimoto et~al.(2018)]{Hashimoto2018ArtificialII}
Hashimoto~DA, Rosman~G, Rus~D, Meireles~OR. 2018. Artificial intelligence in
  surgery: Promises and perils. Annals of Surgery. 268:70–76.

\bibitem[Hersh et~al.(2021)]{Hersh2021AugmentedRI}
Hersh~AM, Mahapatra~S, Weber-Levine~C, Awosika~T, Theodore~JN, Zakaria~HM,
  Liu~A, Witham~TF, Theodore~N. 2021. Augmented reality in spine surgery: A
  narrative review. HSS Journal{\textregistered}: The Musculoskeletal Journal
  of Hospital for Special Surgery. 17:351 -- 358.

\bibitem[Jensen et~al.(2017)]{b824283769d743aeba29c2d3468d1184}
Jensen~S, Wilm~J, Aan{\ae}s~H. 2017. An error analysis of structured light
  scanning of biological tissue. In: Image Analysis; Germany. Springer. p.
  135--145. Scandinavian Conference on Image Analysis, SCIA ; Conference date:
  12-06-2017 Through 14-06-2017.

\bibitem[Kazanzides et~al.(2014)]{kazanzides2014open}
Kazanzides~P, Chen~Z, Deguet~A, Fischer~GS, Taylor~RH, DiMaio~SP. 2014. An
  open-source research kit for the da vinci{\textregistered} surgical system.
  In: 2014 IEEE international conference on robotics and automation (ICRA).
  IEEE. p. 6434--6439.

\bibitem[Kendall et~al.(2018)]{Kendall2018MultitaskLU}
Kendall~A, Gal~Y, Cipolla~R. 2018. Multi-task learning using uncertainty to
  weigh losses for scene geometry and semantics. 2018 IEEE/CVF Conference on
  Computer Vision and Pattern Recognition:7482--7491.

\bibitem[Kingma and Ba(2015)]{Kingma2015AdamAM}
Kingma~DP, Ba~J. 2015. Adam: A method for stochastic optimization. CoRR.
  abs/1412.6980.

\bibitem[Laga et~al.(2020)]{Laga2020ASO}
Laga~H, Jospin~LV, Boussa{\"i}d~F, Bennamoun. 2020. A survey on deep learning
  techniques for stereo-based depth estimation. IEEE transactions on pattern
  analysis and machine intelligence. PP.

\bibitem[Lipson et~al.(2021)]{Lipson2021RAFTStereoMR}
Lipson~L, Teed~Z, Deng~J. 2021. Raft-stereo: Multilevel recurrent field
  transforms for stereo matching. 2021 International Conference on 3D Vision
  (3DV):218--227.

\bibitem[Liu et~al.(2019)]{liu2019end}
Liu~S, Johns~E, Davison~AJ. 2019. End-to-end multi-task learning with
  attention. In: Proceedings of the IEEE/CVF conference on computer vision and
  pattern recognition. p. 1871--1880.

\bibitem[Paszke et~al.(2019)]{NEURIPS2019_9015}
Paszke~A, Gross~S, Massa~F, Lerer~A, Bradbury~J, Chanan~G, Killeen~T, Lin~Z,
  Gimelshein~N, Antiga~L, et~al. 2019. Pytorch: An imperative style,
  high-performance deep learning library. In: Advances in neural information
  processing systems 32. Curran Associates, Inc.; p. 8024--8035.
  \urlprefix\url{http://papers.neurips.cc/paper/9015-pytorch-an-imperative-style-high-performance-deep-learning-library.pdf}.

\bibitem[Qi and Luo(2020)]{Qi2020SmallDC}
Qi~GJ, Luo~J. 2020. Small data challenges in big data era: A survey of recent
  progress on unsupervised and semi-supervised methods. IEEE transactions on
  pattern analysis and machine intelligence. PP.

\bibitem[Rachakonda et~al.(2019)]{Rachakonda2019SourcesOE}
Rachakonda~PK, Muralikrishnan~B, Sawyer~D. 2019. Sources of errors in
  structured light 3d scanners. In: Defense + Commercial Sensing.

\bibitem[Ronneberger et~al.(2015)]{Ronneberger2015UNetCN}
Ronneberger~O, Fischer~P, Brox~T. 2015. U-net: Convolutional networks for
  biomedical image segmentation. In: MICCAI.

\bibitem[Salvi et~al.(2004)]{Salvi2004PatternCS}
Salvi~J, Pag{\`e}s~J, Batlle~J. 2004. Pattern codification strategies in
  structured light systems. Pattern Recognit. 37:827--849.

\bibitem[Scharstein et~al.(2014)]{Scharstein2014HighResolutionSD}
Scharstein~D, Hirschm{\"u}ller~H, Kitajima~Y, Krathwohl~G, Nesic~N, Wang~X,
  Westling~P. 2014. High-resolution stereo datasets with subpixel-accurate
  ground truth. In: GCPR.

\bibitem[Scharstein and
  Szeliski(2003{\natexlab{a}})]{Scharstein2003HighaccuracySD}
Scharstein~D, Szeliski~R. 2003{\natexlab{a}}. High-accuracy stereo depth maps
  using structured light. 2003 IEEE Computer Society Conference on Computer
  Vision and Pattern Recognition, 2003 Proceedings. 1:I--I.

\bibitem[Scharstein and Szeliski(2003{\natexlab{b}})]{1211354}
Scharstein~D, Szeliski~R. 2003{\natexlab{b}}. High-accuracy stereo depth maps
  using structured light. In: 2003 IEEE Computer Society Conference on Computer
  Vision and Pattern Recognition, 2003. Proceedings.; vol.~1. p. I--I.

\bibitem[Schell(2021)]{Ocupe}
Schell~J. 2021. Projectors; [\url{https://github.com/Ocupe/Projectors}].

\bibitem[Sener and Koltun(2018)]{sener2018multi}
Sener~O, Koltun~V. 2018. Multi-task learning as multi-objective optimization.
  Advances in neural information processing systems. 31.

\bibitem[Silberman and Fergus(2011)]{Silberman2011IndoorSS}
Silberman~N, Fergus~R. 2011. Indoor scene segmentation using a structured light
  sensor. 2011 IEEE International Conference on Computer Vision Workshops (ICCV
  Workshops):601--608.

\bibitem[Standley et~al.(2020)]{Standley2020WhichTS}
Standley~TS, Zamir~AR, Chen~D, Guibas~LJ, Malik~J, Savarese~S. 2020. Which
  tasks should be learned together in multi-task learning? In: ICML.

\bibitem[Uhrig et~al.(2017)]{Uhrig2017THREEDV}
Uhrig~J, Schneider~N, Schneider~L, Franke~U, Brox~T, Geiger~A. 2017. Sparsity
  invariant cnns. In: International Conference on 3D Vision (3DV).

\bibitem[Willemink et~al.(2020)]{Willemink2020PreparingMI}
Willemink~MJ, Koszek~WA, Hardell~C, Wu~J, Fleischmann~D, Harvey~H, Folio~LR,
  Summers~RM, Rubin~D, Lungren~MP. 2020. Preparing medical imaging data for
  machine learning. Radiology:192224.

\bibitem[Xu et~al.(2022)]{threephase}
Xu~C, Huang~B, Elson~DS. 2022. Self-supervised monocular depth estimation with
  3-d displacement module for laparoscopic images. IEEE Transactions on Medical
  Robotics and Bionics. 4(2):331--334.

\bibitem[Ying(2019)]{Ying2019AnOO}
Ying~X. 2019. An overview of overfitting and its solutions. Journal of Physics:
  Conference Series.

\bibitem[Zhao et~al.(2020)]{Zhao2020MonocularDE}
Zhao~C, Sun~Q, Zhang~C, Tang~Y, Qian~F. 2020. Monocular depth estimation based
  on deep learning: An overview. Science China Technological Sciences:1--16.

\end{thebibliography}

\end{document}